\documentclass[letterpaper, 10 pt, conference]{ieeeconf}  

\IEEEoverridecommandlockouts                              

\usepackage{graphics} 
\usepackage{epsfig} 
\usepackage{amsmath} 
\usepackage{amssymb}  
\usepackage{amsfonts}
\usepackage{commath}
\usepackage{cleveref}
\usepackage{graphicx}
\usepackage[rgb]{xcolor}
\usepackage{booktabs}
\usepackage{multirow}

\crefname{equation}{}{} 

\title{\LARGE \bf
Enhancing Model-Based Step Adaptation for Push Recovery through Reinforcement Learning of Step Timing and Region
}

\author{Tobias Egle*$^{,1}$, Yashuai Yan*$^{,2}$, Dongheui Lee$^{2,3}$, and Christian Ott$^{1,3}$
	\thanks{$^{1}$Tobias Egle and Christian Ott are with the Automation and Control Institute, TU Wien, 1040 Vienna, Austria
		{\tt\small \{tobias.egle, christian.ott\}@tuwien.ac.at}}%
	\thanks{$^{2}$Yashuai Yan and Dongheui Lee are with the Autonomous Systems Lab, TU Wien, 1040 Vienna, Austria
		{\tt\small \{yashuai.yan, dongheui.lee\}@tuwien.ac.at}}%
  \thanks{$^{3}$Christian Ott and Dongheui Lee are also with the Institute of Robotics and Mechatronics (DLR), German Aerospace Center, Wessling, Germany.}%
	\thanks{This project has received funding from the European Research Council (ERC) under the European Union’s Horizon 2020 research and innovation programme (grant agreement No. 819358).}%
    \thanks{*Contributed equally to the work.}
}

\begin{document}

\maketitle
\thispagestyle{empty}
\pagestyle{empty}

\begin{abstract}
This paper introduces a new approach to enhance the robustness of humanoid walking under strong perturbations, such as substantial pushes. Effective recovery from external disturbances requires bipedal robots to dynamically adjust their stepping strategies, including footstep positions and timing. Unlike most advanced walking controllers that restrict footstep locations to a predefined convex region, substantially limiting recoverable disturbances, our method leverages reinforcement learning to dynamically adjust the permissible footstep region, expanding it to a larger, effectively non-convex area and allowing cross-over stepping, which is crucial for counteracting large lateral pushes. Additionally, our method adapts footstep timing in real time to further extend the range of recoverable disturbances. Based on these adjustments, feasible footstep positions and DCM trajectory are planned by solving a QP. Finally, we employ a DCM controller and an inverse dynamics whole-body control framework to ensure the robot effectively follows the trajectory.
\end{abstract}

\section{INTRODUCTION}
In humanoid robot locomotion, Divergent Component of Motion (DCM) trajectory generation \cite{takenakaRealTimeMotion2009a} has become a widely used method for generating walking motions. Current research is evolving from focusing on motion generation to addressing the challenges of robustness and adaptability in the face of external disturbances. Model-based push recovery approaches often involve analytical calculation of footstep adjustments \cite{englsbergerThreeDimensionalBipedalWalking2015,mesesanOnlineDCMTrajectory2021} or optimization-based methods \cite{Khazoom_2024} to adjust timing and position of footsteps \cite{khadivWalkingControlBased2020,egleStepTimingAdaptation2023}. Data-driven approaches have also been applied to generate push recovery strategies. Recent works \cite{yangLearningWholebodyMotor2018,ferigoEmergenceWholeBodyStrategies2021} used end-to-end Reinforcement Learning (RL) to acquire a variety of push recovery and balancing behaviors, such as ankle, hip, and stepping strategies, which are similar to those seen in humans \cite{horakCentralProgrammingPostural1986}. While these methods adapt well to various disturbances, they often require extensive training data and significant computational resources.

Model-based approaches typically use distinct control components, such as a low-level whole-body controller and a high-level trajectory planner to precisely follow pre-planned motions \cite{englsbergerTorqueBasedDynamicWalking2018a}. In contrast, data-driven approaches often employ end-to-end learning of joint positions \cite{singhLearningBipedalWalking2023,xieALLSTEPSCurriculumdrivenLearning2020a,rudinLearningWalkMinutes2022,liReinforcementLearningVersatile2024}. These methods tend to generalize better to unknown environments due to extensive domain randomization. For instance, \cite{siekmannBlindBipedalStair2021} showcased robust blind walking on stairs using terrain randomization techniques. Moreover, \cite{duanLearningDynamicBipedal2022} demonstrates navigation over difficult stepping-stone patterns effectively using vision input. Nevertheless, the success of end-to-end learning approaches largely depends on the precise design of the reward function.
\begin{figure}
    \centering
    \includegraphics[width=\linewidth]{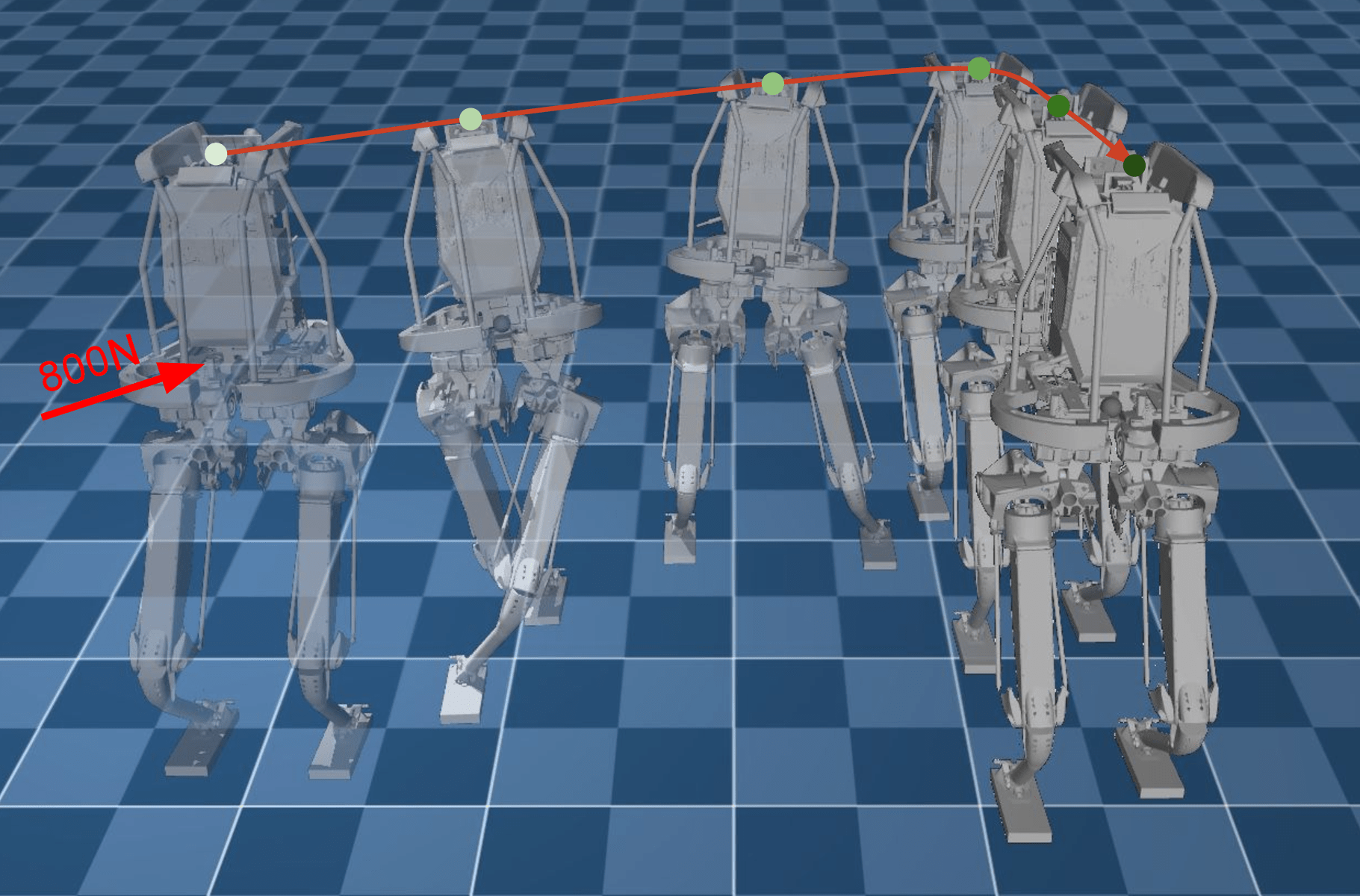}
    \caption{Simulation of the robot kangaroo during a lateral push force of 800N for 0.1 seconds. Our method allows the robot to quickly recover from such a large disturbance through a leg cross-over and simultaneous adjustment in step timing.}
    \label{fig:enter-label}
    \vspace{-15pt}
\end{figure}%

Hybrid approaches that combine model-based and data-driven methods have emerged as promising solutions, aiming to take advantage of both \cite{baoDeepReinforcementLearning2024, gu2023walkingbylogicsignaltemporallogicguided}. Duan et al. \cite{duanLearningTaskSpace2021} propose integrating robot system knowledge into reinforcement learning to train bipedal locomotion policies directly in task space, improving sample efficiency and demonstrating the approach in simulation and on the real robot Cassie. Castillo et al. \cite{castilloReinforcementLearningBasedCascade2022} achieved a lightweight network structure and sample efficiency through an RL framework, where actions parameterize desired joint trajectories rather than direct actuator inputs. In \cite{castilloTemplateModelInspired2023b}, the same authors propose a hierarchical approach that uses RL at the high level to train policies for task space commands and employs a model-based low-level controller to track these trajectories. Both policies demonstrate robust performance against various disturbance forces applied to the torso.

From a model-based perspective, we can effectively find optimal solutions for previewed footstep placements using a QP approach. However, some drawbacks remain. The step time appears nonlinear in the solution of the DCM dynamics. Thus, iterative solutions or a general nonlinear optimization are required. Additionally, using a QP restricts possible footstep regions to convex shapes, thereby reducing the solution space and excluding options such as cross-over stepping. Recent works have addressed this problem by decomposing the non-convex regions into convex subregions by a set of rules \cite{griffinReachabilityAwareCapture2023} or by evaluating the feasibility of the resulting QP problem \cite{habibHandlingNonConvexConstraints2022}.

The contribution of this work lies in enhancing model-based trajectory planning by integrating RL to address the limitations of the QP approach. This is achieved by using an RL agent to dynamically adjust key parameters within the model-based control framework. The selected parameters are the step frequency, single support percentage, and rotation angle of a convex step area around the current stance foot. Thanks to the combination of RL with the model-based framework, our method significantly improves the maximal recoverable external disturbances, enhancing the robustness of the robot's walking. Additionally, the division of tasks between RL and model-based controllers improves the learning efficiency of RL, enabling the training of a single environment in just a few hours.


\section{Preliminaries}\label{sec:prel}

Our model-based walking trajectory generation uses the concepts of the three-dimensional DCM and the Virtual Repellent Point (VRP) \cite{englsbergerThreeDimensionalBipedalWalking2015}. The DCM is defined as
\begin{equation}
    \boldsymbol{\xi} = \boldsymbol{x} + b \dot{\boldsymbol{x}},
\end{equation}
where $\boldsymbol{x}$ is the Center of Mass (CoM) and $b = \sqrt{\Delta z / g}$ is given by the average CoM height $\Delta z$ and gravity $g$. From the CoM dynamics $\ddot{\boldsymbol{x}} = \boldsymbol{g} + \boldsymbol{F}_\mathrm{ext}/m$ the DCM dynamics can be derived as
\begin{equation}\label{eq:DCMdynamics}
    \dot{\boldsymbol{\xi}} = \frac{1}{b} (\boldsymbol{\xi} - \boldsymbol{v}).
\end{equation}
Here, the VRP $\boldsymbol{v}$ encodes the total force $\boldsymbol{F}=\boldsymbol{F}_g + \boldsymbol{F}_\mathrm{ext}$ on the CoM as $\boldsymbol{F} = \frac{m}{b^2} (\boldsymbol{x} - \boldsymbol{v})$, where $m$ is the total mass.
\subsection{DCM trajectory generation (backward recursion)}\label{sec:DCMtraj}
The trajectory is divided into $n_\mathrm{\varphi}$ transition phases (alternating single and double support). Within each transition phase $\varphi$ with time $t \in [0,T]$ we assume a spatially linear interpolation for the VRP between a set of $n_{\mathrm{wp}} = n_\mathrm{\varphi}+1$ waypoints $\{\boldsymbol{v}_i\}$. Given the VRP trajectory, we can solve \cref{eq:DCMdynamics} with a DCM terminal constraint $\boldsymbol{\xi}_{\varphi,T}$ as
 \begin{equation}\label{eq:SolDCMDyn}
		\boldsymbol{\xi}_{\varphi}(t)=\alpha_{\varphi}(t) \boldsymbol{v}_{\varphi,0}+\beta_{\varphi}(t) \boldsymbol{v}_{\varphi,T}+\gamma_{\varphi}(t) \boldsymbol{\xi}_{\varphi,T},
\end{equation}
where $\alpha_{\varphi}(t)$, $\beta_{\varphi}(t)$, and $\gamma_{\varphi}(t)$ are nonlinear coefficients in time that depend on the temporal interpolation between the VRP start point $\boldsymbol{v}_{\varphi,0}$ and endpoint $\boldsymbol{v}_{\varphi,T}$ (cf. \cite{mesesanConvexPropertiesCenterofMass2018}). The start and end points of adjacent transition phases are linked to ensure trajectory continuity. Starting from a DCM endpoint, this calculation is performed backward in time, hence the term "backward recursion".
\subsection{DCM tracking control}
\label{sec:DCMtrackingControl}
To control the unstable first-order DCM dynamics, Englsberger et al. \cite{englsbergerThreeDimensionalBipedalWalking2015} proposed tracking the reference DCM trajectory using the following control law:

\begin{equation}\label{eq:DCMcontroller}
\boldsymbol{v} = \boldsymbol{v}_{\mathrm{ref}} + (\boldsymbol{I} + b \boldsymbol{K}_{\xi}) (\boldsymbol{\xi} - \boldsymbol{\xi}_{\mathrm{ref}}),
\end{equation}

where \( \boldsymbol{v}_{\mathrm{ref}} \) is the reference VRP position and \( \tilde{\boldsymbol{\xi}} = \boldsymbol{\xi} - \boldsymbol{\xi}_{\mathrm{ref}} \) is the DCM tracking error. This approach ensures stable tracking for a positive diagonal matrix \( \boldsymbol{K}_{\xi} \).
\subsection{Separation into Ankle and Step Strategy}\label{sec:AnkleStep}
Mesesan et al. \cite{mesesanOnlineDCMTrajectory2021} proposed splitting the current DCM error \( \tilde{\boldsymbol{\xi}}\) into two parts:
\begin{equation}
\tilde{\boldsymbol{\xi}} = \tilde{\boldsymbol{\xi}}_{\text{ankle}} + \tilde{\boldsymbol{\xi}}_{\text{step}},
\end{equation}
where \( \tilde{\boldsymbol{\xi}}_{\text{ankle}} \) is correctable by the ankle strategy, and \( \tilde{\boldsymbol{\xi}}_{\text{step}} \) requires the step strategy. The ankle strategy, i.e., the DCM controller \cref{eq:DCMcontroller}, can use VRP adjustments that remain in the support polygon. For example, the set of possible VRP adjustments usable by the ankle strategy is defined as \( \tilde{V} = \{ \tilde{v} = (x, y, 0)^T \mid -l \leq x \leq l, -w \leq y \leq w \} \) for a rectangular foot with a width of $2w$ and a length of $2l$. Inserting this set of VRP adjustments in the recursive solution of the DCM dynamics \cref{eq:SolDCMDyn}, we can compute the set of correctable DCM errors in the first phase as
\begin{equation}
\tilde{\Xi}_{\text{ankle}} = \alpha_{1}(t) \tilde{V}_1 + \beta_{1}(t) \tilde{V}_2 + \gamma_{1}(t) \tilde{\Xi}_2.
\end{equation}
Here, $\tilde{\Xi}_2$ is the remaining error at the end of the phase. This computation can be extended to multiple transition phases. Projecting the current DCM error \( \tilde{\boldsymbol{\xi}} \) onto \( \tilde{\Xi}_{\text{ankle}} \) minimizes \( \lVert \tilde{\boldsymbol{\xi}}_{\text{step}} \rVert \), ensuring the step strategy activates only if \( \tilde{\boldsymbol{\xi}} \) is outside \( \tilde{\Xi}_{\text{ankle}} \). A graphical illustration of this computation is indicated in Fig. \ref{fig:FootRegionRotation}-2). For more details, see \cite{mesesanOnlineDCMTrajectory2021,egleStepTimingAdaptation2023}.
\section{Methodology}\label{sec:method}
\begin{figure*}[]
    \vspace{5pt}
    \centering
    \includegraphics[width=0.99\textwidth]{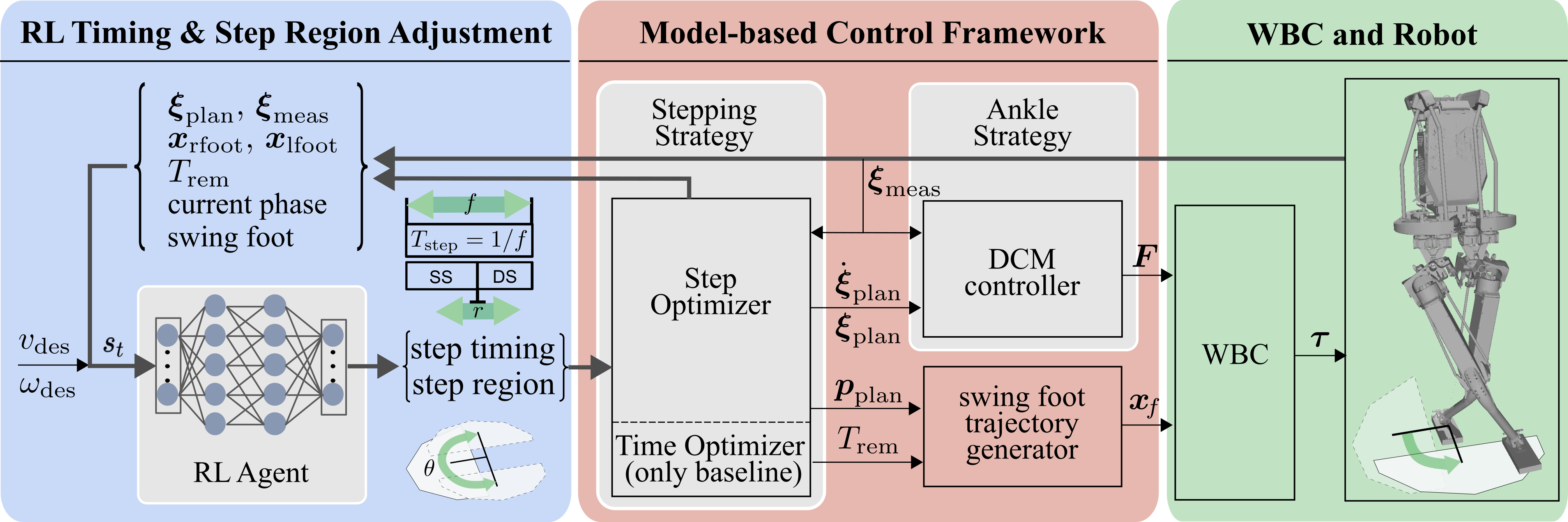}
    \caption{\textbf{Overview of our framework.} The main contribution is extending the model-based control framework by an RL-based step timing and region adaptation. An inverse dynamics whole-body controller generates the desired joint torques. Model-based time adaptation is only used in the baseline method.}
    \label{fig:overview}
\end{figure*}


\subsection{Overview of our framework}
An overview of our walking control framework is illustrated in Fig. \ref{fig:overview}. The architecture mainly consists of two components: the RL for step timing and region adaptation and the model-based control framework. While we utilize an inverse dynamics Whole-Body Controller (WBC) \cite{englsbergerTorqueBasedDynamicWalking2018a} in the final component, our contribution does not focus on this aspect.
The RL framework is used to dynamically adjust selected parameters that influence step timing and step region. Under these adjustments, the model-based method plans the foot and DCM trajectories. The WBC optimizes the joint torques to follow the desired trajectories.

\subsection{Model-based step timing adaptation}
\label{sec:dcm}
While our approach is based on the model-based trajectory generation outlined in \cite{egleStepTimingAdaptation2023}, we introduce several modifications to enhance the previous framework. By directly choosing the DCM and VRP waypoints as optimization variables, finding an exact solution for the time adjustment during the single support phase is possible. However, due to the interpolation of the VRP between two foot positions in the double support phase, we still rely on an iterative solution during this phase (details in Sec. \ref{sec:time_adjustment}). Finally, different from \cite{egleStepTimingAdaptation2023}, we only minimize the distance of the final DCM from its nominal value, which is sufficient for convergence of the trajectory.

\begin{figure}
    \vspace{5pt}
    \centering
    \includegraphics[width=0.48\textwidth]{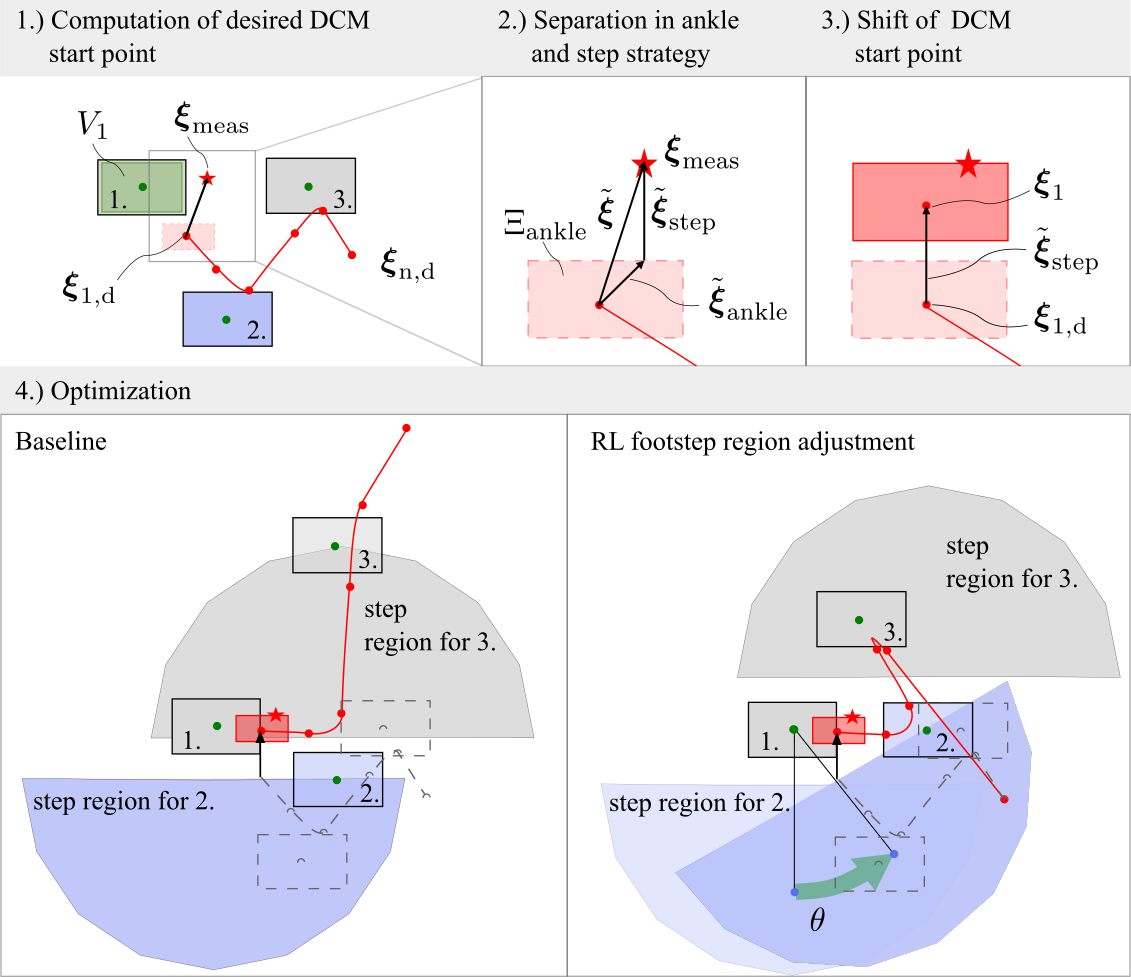}
    \caption{Graphical illustration of the computation of the DCM start point as input to the optimization problem. Comparison of model-based step timing adaptation and improvement due to possible adjustment of the footstep region by the RL agent.}
\label{fig:FootRegionRotation}
\end{figure}

A graphical overview of the DCM trajectory generation is shown in Fig. \ref{fig:FootRegionRotation}. The backward computation approach introduced in Section \ref{sec:DCMtraj} is well-suited for offline trajectory generation. The DCM terminal constraint guarantees convergence of the trajectory. However, with predefined VRP waypoints, the initial DCM cannot be controlled. Therefore, the measured DCM must be close to the trajectory to be feasible. This proximity ensures that the ankle strategy produces feasible VRP adjustments according to \cref{eq:DCMcontroller}.

In online trajectory generation, we continuously replan the trajectory and only evaluate the next time step of the trajectory. In this context, the initial DCM state holds greater importance than strict convergence to a predefined endpoint. We will address this convergence using a soft constraint in the subsequent optimization problem.
\subsubsection{DCM trajectory generation as an optimization problem}
By solving the DCM dynamics \cref{eq:DCMdynamics} and evaluating it at the end time $T$ of a transition phase, we obtain the recursive equation for the DCM waypoints as
\begin{equation}\label{eq:forwardDCM}
	\begin{aligned}
		\boldsymbol{\xi}_{i}&=\underbrace{\left(-\frac{b}{T}-e^{\frac{T}{b}}\left(1-\frac{b}{T}\right) \right)}_{\alpha_{T,i-1}} \boldsymbol{v}_{i-1}+\underbrace{\left(1+\frac{b}{T}- e^{\frac{T}{b}}\frac{b}{T}\right)}_{\beta_{T,i-1}} \boldsymbol{v}_{i}\\&+\underbrace{e^{\frac{T}{b}}}_{\gamma_{T,i-1}} \boldsymbol{\xi}_{i-1} \quad \forall i \in \{2,\dots,n_{\mathrm{wp}}\}.
	\end{aligned}
\end{equation}
Here, unlike \cref{eq:SolDCMDyn}, we start from a DCM start point and compute the next waypoint in a "forward" recursion. To reduce the size of the optimization problem, we choose to express the VRP waypoints in terms of foot positions $\boldsymbol{p}_j$ as
\begin{equation}\label{eq:foot2v}
		\boldsymbol{v}_i = \boldsymbol{p}_j + [0 \enspace 0 \enspace \Delta z]^{T}
		\enspace \text{with} \enspace
		j = \begin{cases}\lceil i/2 \rceil\enspace\text{for cSS}\\
			\lfloor i/2 + 1 \rfloor\enspace \text{for cDS}.
		\end{cases}
\end{equation}%
for all $i \in \{1,\dots,n_{\mathrm{wp}}\}$, for all $j \in \{1,\dots,n_{\mathrm{pfs}}\}$ and obtain
\begin{equation}\label{eq:DCMwayp}
        \boldsymbol{\xi}_{i} =\begin{cases}\left(\alpha_{T,i-1}+\beta_{T,i-1}\right) \boldsymbol{p}_{j}+\gamma_{T,i-1} \boldsymbol{\xi}_{i-1}\enspace\text{for SS}\\
        \alpha_{T,i-1} \boldsymbol{p}_{j}+\beta_{T,i-1} \boldsymbol{p}_{j+1}+\gamma_{T,i-1} \boldsymbol{\xi}_{i-1}\enspace\text{for DS}.
		\end{cases}
\end{equation}
Here, $n_{\mathrm{pfs}}$ is the number of foot positions, and the terms "cSS" and "cDS" in \cref{eq:foot2v} denote the current ongoing phases of support, whereas "SS" and "DS" in \cref{eq:DCMwayp} refer to the general phases of single and double support, respectively.
To easily relate the trajectory generation to the current state of the robot, we start the preview trajectory from the current time in the transition phase $t_1$, i.e., $\boldsymbol{\xi}_1 = \boldsymbol{\xi}(t_1)$ and $v_1 = v(t_1)$. Thus, the duration of the first phase is the remaining time $T_\mathrm{rem}$ and not the total time of the transition phase $T_1$. In the single support phase, the start and end VRP are equal to the foot position, whereas, in the double support phase, the VRP is interpolated between two feet positions (see \cref{eq:DCMwayp}). Consequently, the solution for the DCM in the current transition phase differs from \cref{eq:DCMwayp} as follows
\begin{equation}\label{eq:DCMwayp1}
        \boldsymbol{\xi}_{2} =\begin{cases} \left(
        \alpha_{\, T_\mathrm{rem}} + \beta_{\, T_\mathrm{rem}}\right) \boldsymbol{p}_{j}+\gamma_{\, T_\mathrm{rem}} \boldsymbol{\xi}_{1}\enspace\text{for cSS}\\
        \alpha_{\, T_\mathrm{rem}} \boldsymbol{v}_{1}+\beta_{\, T_\mathrm{rem}} \boldsymbol{p}_{j+1}+\gamma_{\, T_\mathrm{rem}} \boldsymbol{\xi}_{1}\enspace\text{for cDS}.
        \end{cases}
\end{equation}
\subsubsection{Time adjustment in first transition phase} \label{sec:time_adjustment}
Since in the initial transition phase, the foot positions do not change, and $\boldsymbol{\xi}_{1}$ is a known input parameter, we can use this phase for time adjustment. Here, we have to differentiate between single and double support phases. We omit the subscripts of the coefficients $\alpha$, $\beta$, and  $\gamma$ for readability and use $T$ instead of the correct $T_\mathrm{rem}$ in this section.

\emph{Single support}: With the constraint $\alpha + \beta + \gamma = 1$ (see \cite{mesesanConvexPropertiesCenterofMass2018}), we can formulate \cref{eq:DCMwayp1} for the single support phase as
\begin{equation}\label{eq:DCM1SS}
\boldsymbol{\xi}_{2} = (1-\gamma) \boldsymbol{p}_{j} + \gamma \boldsymbol{\xi}_1.
\end{equation}
Here, we can directly solve for $\gamma$ in the optimization problem and substitute $T^\ast = b \ln(\gamma)$ to obtain the optimal transition phase time.

\emph{Double support}: Due to the interpolation of the VRP between two feet positions, we cannot find an exact solution for the transition phase time and thus need to linearize the coefficients $\alpha$, $\beta$, and  $\gamma$ around a nominal value $T_0$. According to \cref{eq:DCMwayp1} we obtain the next DCM waypoint in the first phase in double support as
\begin{equation}
\boldsymbol{\xi}_{2} = \alpha \boldsymbol{v}_1 + \beta \boldsymbol{p}_{j+1} + \gamma \boldsymbol{\xi}_1,
\end{equation}
with
\begin{equation}
\gamma = e^{T/b}, \quad \beta = \frac{b}{T} + 1 - \gamma \frac{b}{T}, \quad \alpha = 1 - \beta - \gamma.
\end{equation}
The linearized forms of the coefficients are:
\begin{equation}
\gamma \approx \gamma_{\text{c}} + \gamma_{\ell} T, \quad \beta \approx \beta_{\text{c}} + \beta_{\ell} T, \quad \alpha \approx \alpha_{\text{c}} + \alpha_{\ell} T,
\end{equation}
where
\begin{equation}
\gamma_{\text{c}} = e^{T_0/b}, \quad \gamma_{\ell} = \frac{e^{T_0/b}}{b},
\end{equation}
\begin{equation}
\beta_{\text{c}} = \frac{b}{T_0} + 1 - \gamma_{\text{c}} \frac{b}{T_0}, \quad \beta_{\ell} = -\frac{b}{T_0^2} + \frac{\gamma_{\text{c}} b}{T_0^2} - \frac{\gamma_{\ell} b}{T_0},
\end{equation}
\begin{equation}
\alpha_{\text{c}} = 1 - \beta_{\text{c}} - \gamma_{\text{c}}, \quad \alpha_{\ell} = -\beta_{\ell} - \gamma_{\ell}.
\end{equation}
The equation in the first phase can be rewritten as
\begin{equation}\label{eq:DCMLinConst}
\boldsymbol{\xi}_{2} \approx (\alpha_{\ell} \boldsymbol{v}_1 + \beta_{\ell} \boldsymbol{p}_{j+1} + \gamma_{\ell} \boldsymbol{\xi}_1) T + (\alpha_{\text{c}} \boldsymbol{v}_1 + \beta_{\text{c}} \boldsymbol{p}_{j+1} + \gamma_{\text{c}} \boldsymbol{\xi}_1).
\end{equation}
In the double support phase, we must limit the time adjustments to ensure that the linearization remains a valid approximation.

\begin{figure}
    \centering
    \includegraphics[width=0.33\textwidth]{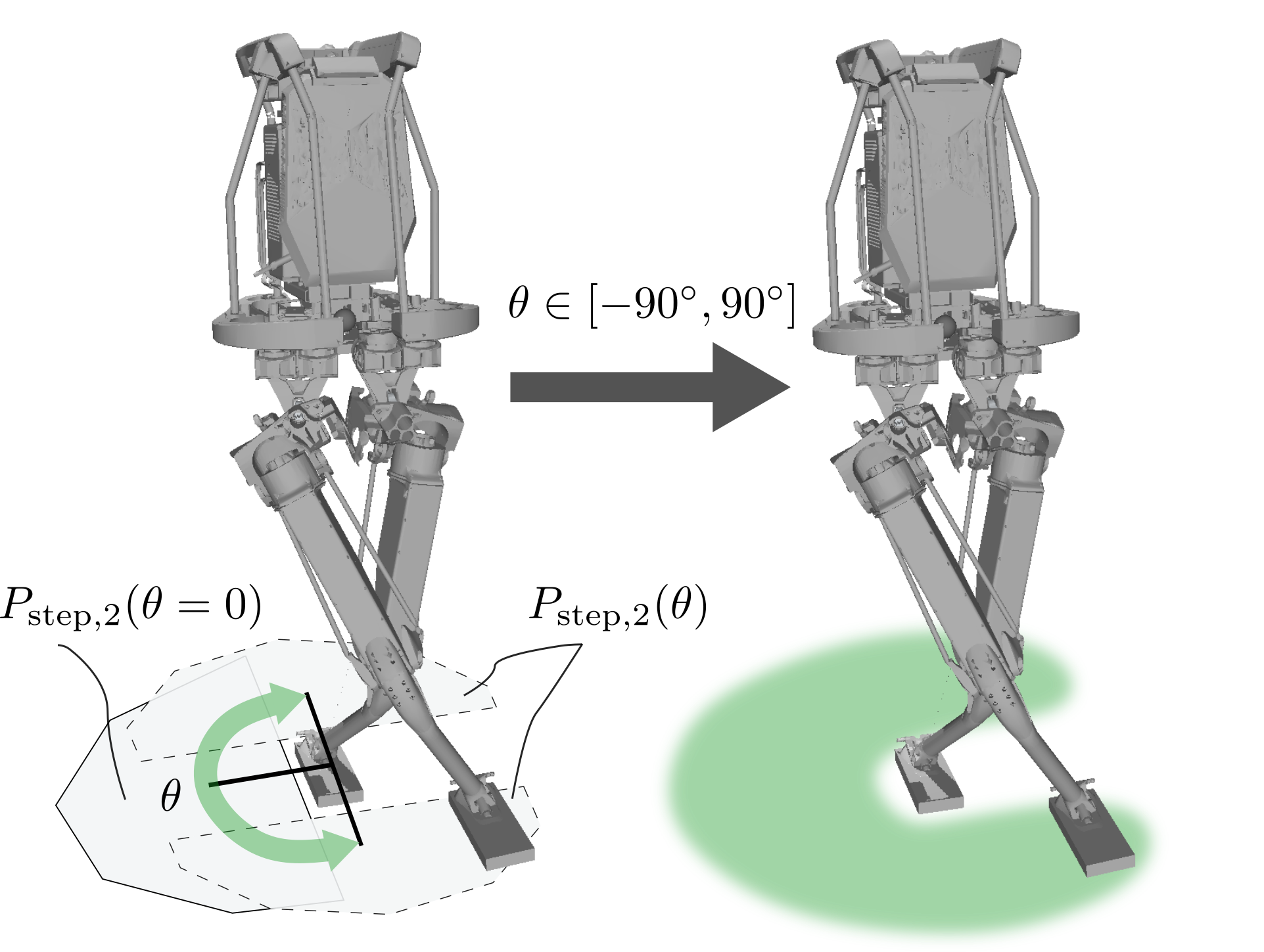}
    \caption{\textbf{Step region parametrization.} The extended footstep region is parametrized by a convex region (here $P_{\mathrm{step},2}$ for single support) and a rotation angle $\theta$ around the stance foot. }
    \label{fig:step_rotation}
\end{figure}
\subsubsection{DCM initial condition}
Since the DCM trajectory is calculated using a forward recursion according to \cref{eq:forwardDCM}, the starting point for the DCM can be freely chosen. However, due to the divergent nature of DCM dynamics, selecting the starting point requires careful consideration. Based on the current nominal footstep plan, we choose a DCM terminal condition $\boldsymbol{\xi}_{n,\mathrm{d}}$ between the last two footstep positions, as illustrated in Fig. \ref{fig:FootRegionRotation}. We then calculate the DCM trajectory backward and obtain the DCM starting point $\boldsymbol{\xi}_{1,\mathrm{d}}$, from which we can follow the DCM trajectory under ideal conditions. To stabilize this trajectory, we first define the DCM error as $\tilde{\boldsymbol{\xi}} = \boldsymbol{\xi} -\boldsymbol{\xi}_{1,\mathrm{d}}$. We aim to integrate both ankle and step strategies for maximal robustness against disturbances by dividing the DCM error into two components, i.e., $\tilde{\boldsymbol{\xi}} = \tilde{\boldsymbol{\xi}}_\mathrm{ankle} + \tilde{\boldsymbol{\xi}}_\mathrm{step}$. We do this by projecting the current DCM error onto a set of DCM errors that can be corrected using the ankle strategy, thereby minimizing the error that needs to be compensated for by adjusting the reference trajectory, i.e., through step adjustments (cf. \cite{mesesanOnlineDCMTrajectory2021}). We add the DCM error for step adjustment to the desired DCM start point to obtain the DCM start point as
\begin{equation}
\vspace{-10pt}
    \boldsymbol{\xi}_1 = \boldsymbol{\xi}_{1,\mathrm{d}} + \tilde{\boldsymbol{\xi}}_\mathrm{step}.
\label{eq:dcm_error_decompose}
\vspace{-10pt}
\end{equation}
\subsubsection{Step and Timing Optimization}
We cannot follow the nominal DCM trajectory due to the shifted DCM start point. Instead, we formulate an optimization problem that adjusts the time in the first transition phase, all previewed footstep positions, and the DCM endpoint to obtain a feasible DCM trajectory. The optimization problem in the single support phase is formulated as a QP as
\begin{equation}\label{eq:qpSS}
	\begin{aligned}
		\min_{\gamma,\boldsymbol{p}_j,\boldsymbol{\xi}_i} & w_{\gamma} \abs{\gamma-\gamma_{\mathrm{d}}}^2 + \!\! \sum_{j=2}^{2+n_{\mathrm{pfs}}}\norm{\boldsymbol{p}_j-\boldsymbol{p}_{j,\mathrm{d}}}^2_{\boldsymbol{W}_{\negthickspace p}} \! + \norm{\boldsymbol{\xi}_n-\boldsymbol{\xi}_{n,\mathrm{d}}}^2_{\boldsymbol{W}_{\negthickspace \xi}}\\
		\textrm{s.t.} & \quad \text{\cref{eq:DCMwayp,eq:DCM1SS}},\\
        & \quad e^{\frac{T_{\mathrm{min}}}{b}} \leq \gamma \leq e^{\frac{T_{\mathrm{max}}}{b}},\enspace
		\boldsymbol{p}_j \in P_{\mathrm{step},j}.\\
	\end{aligned}
\end{equation}
Similarly, the optimization problem in the double support phase is given by
\begin{equation}\label{eq:qpDS}
	\begin{aligned}
		\min_{T,\boldsymbol{p}_j,\boldsymbol{\xi}_i} & w_{T} \abs{T-T_{0}}^2 +\!\! \sum_{j=3}^{3+n_{\mathrm{pfs}}}\norm{\boldsymbol{p}_j-\boldsymbol{p}_{j,\mathrm{d}}}^2_{\boldsymbol{W}_{\negthickspace p}} \! + \norm{\boldsymbol{\xi}_n-\boldsymbol{\xi}_{n,\mathrm{d}}}^2_{\boldsymbol{W}_{\negthickspace \xi}}\\
		\textrm{s.t.} & \quad \text{\cref{eq:DCMwayp,eq:DCMLinConst}},\\
  & \quad T_\mathrm{min} \leq T \leq T_\mathrm{max}, \enspace \boldsymbol{p}_j \in P_{\mathrm{step},j}.
	\end{aligned}
\end{equation}
With the solution of the optimization problem, we update the reference VRP and DCM as
\begin{equation}\label{eq:currVRP}
	\boldsymbol{v}_1(t_s) =\left(1-\frac{t_s}{T_\mathrm{rem}}\right) \boldsymbol{v}_{1}+\frac{t_s}{T_\mathrm{rem}} \boldsymbol{v}_{2},
\end{equation}
\begin{equation}\label{eq:xiref}
	\boldsymbol{\xi}(t_s)=\alpha_{T_\mathrm{rem}}(t_s) \boldsymbol{v}_1+\beta_{T_\mathrm{rem}}(t_s) \boldsymbol{v}_2+\gamma_{T_\mathrm{rem}}(t_s) \boldsymbol{\xi}_1,
\end{equation}
with $\boldsymbol{v}_{2}$ according to \cref{eq:foot2v} and $t_s$ is the sample time.

\subsection{Reinforcement learning of step time and region}\label{sec:rl}

To enhance the model-based framework, we integrate RL to dynamically adjust key parameters of the trajectory generator. The input to our RL is shown in Fig.~\ref{fig:overview}. The \textit{swing foot} is represented as binary values, i.e., $+1$ for the left foot and $-1$ for the right foot. Similarly, for the \textit{current phase}, $+1$ indicates the single support phase and $-1$ the double support phase.

\subsubsection{Step region adjustment}
\label{sec:step_region}
In model-based trajectory generation, the foot positions are treated as variables that can be optimized in a QP framework, constraining foot placement to a convex region. The convex region is typically defined on the same side as the swing foot, preventing foot collisions but severely restricting foot placement options for robots. 

In this study, we parameterize the footstep region as a convex region and an angle $\theta$ (see Fig. \ref{fig:step_rotation}). The angle $\theta$ is a continuous variable ranging from $-90^{\circ}$ to $90^{\circ}$, rotating the convex region around the current footstep. This creates an effectively non-convex possible footstep region, even though the QP still operates under convex constraints. Our method employs RL to adjust the parameter $\theta$ to expand the permissible step region. Although QP ensures a collision-free swing foot target within the convex footstep region, the trajectory leading to this point may not be collision-free. The RL algorithm learns through significant penalties to avoid self-collisions by only rotating the footstep region when a collision-free path exists to the swing foot target.

\subsubsection{Step timing adjustment}
\label{sec:step_timing}
Unlike the model-based method in Sec. \ref{sec:dcm} that optimizes the remaining transition time only in the first phase, our RL adjusts the step frequency $f$ and single support percentage $r$ for all phases involved in trajectory generation. In this case, the model-based time adaptation corresponding to the first term in the cost function in \cref{eq:qpSS,eq:qpDS} is disabled. The model-based method approximates the nonlinear optimization in the double support phase by linearizing the transition time around the nominal value, further limiting its ability to find optimal step timing.

Our RL-based time adjustment aims to maintain nominal walking behavior whenever possible, with adjustments to step timing made only when necessary. For minor disturbances, if the robot can recover using its ankle strategy via \cref{eq:DCMcontroller}, the step timing remains unchanged. However, the step timing is adjusted when the disturbance is too significant for the ankle strategy alone. This time adjustment can be formulated as follows:

\begin{equation}
\begin{aligned}
    f &= f_{\text{nom}} + \eta \hat{f}, \\
    r &= r_{\text{nom}} + \eta \hat{r},
\end{aligned}
\label{eq:timing}
\end{equation}
where
\begin{equation*}
    \eta = \left\{\begin{matrix}
0 & \widetilde{\xi}_{\text{step}} = 0 \\ 
1 & \text{otherwise}.
\end{matrix}\right.
\end{equation*}
Here, $\widetilde{\xi}_{\text{step}} = 0$ indicates the disturbance can be managed by the ankle strategy (see \cref{eq:dcm_error_decompose}). Otherwise, the RL adjusts the step timing with $\hat{f}$ and $\hat{r}$ accordingly.

\subsubsection{Rewards}
\label{sec:rewards}


\begin{table}[]
\vspace{5pt}
\centering
\caption{Reward functions.}
\begin{tabular}{lccc}
Reward &      Expression             & Distance/Condition & Parameter \\ \midrule
$R_{\text{freq}}$ & \multirow{4}{*}{$\omega e^{-\lambda d}$} & $d = |\hat{f}|$ & $\omega = 2.0; \lambda = 1$ \\
$R_{\text{ss}}$ &                   & $d = |\hat{r}|$ & $\omega = 1.0; \lambda = 5$  \\
$R_{\text{reg}}$ &                   & $d = |\theta|$ & $\omega = 0.1; \lambda = 2$ \\
$R_{\text{dcm}}$ &                   & $d = \norm{\boldsymbol{\xi}_n-\boldsymbol{\xi}_{n,\mathrm{d}}}$ & $\omega = 10; \lambda = 1$ \\ \midrule
$R_{\text{fall}}$ & \multirow{2}{*}{$\left\{\begin{matrix} -\lambda & \text{Condition}\\ 0 & \text{otherwise} \end{matrix}\right.$} & robot falls & $\lambda = 200$  \\
$R_{\text{colli}}$ &            & self-collision & $\lambda = 200$  \\ \midrule
\end{tabular}
\label{tab:reward}
\vspace{-10pt}
\end{table}
In our method, RL is used to adjust high-level footstep regions and timing, while precise foot locations are optimized via a QP approach under these adjustments. This combination of RL and a model-based framework significantly reduces the effort required to design reward functions. The applied reward functions are shown in Table~\ref{tab:reward}.

The rewards $R_{\text{freq}}, R_{\text{ss}}$, and $R_{\text{reg}}$ encourage the step timing and region to remain close to their nominal values if possible. Similarly to the DCM cost in optimization in \cref{eq:qpSS}, we employ a DCM reward $R_{\text{dcm}}$ to penalize the distance between the final DCM waypoint $\boldsymbol{\xi}_n$ and the final desired DCM waypoint $\boldsymbol{\xi}_{n,d}$. Finally, we penalize if the robot falls or collides with itself in rewards $R_{\text{fall}}, R_{\text{colli}}$.

The parameters $\omega$ and $\lambda$ are hyperparameters for training. The $\lambda$ in the exponential function serves to normalize the distances to facilitate later distributing the importance by $\omega$ over different rewards. However, fine-tuning these parameters is not the main focus of this paper. Better results can be achieved with further fine-tuning. Our total reward is a sum of these rewards:
\begin{equation}
    R_{\text{total}} = R_{\text{freq}} + R_{\text{ss}} + R_{\text{reg}} + R_{\text{dcm}} + R_{\text{fall}} + R_{\text{colli}}.
\label{eq:rew}
\end{equation}

\section{Experiments}\label{sec:exp}
\subsection{Experiment setting}
For RL, the PPO algorithm \cite{schulmanProximalPolicyOptimization2017} is utilized to learn footstep timing and region adjustments. The actor and critic networks are designed as multi-layer perceptions (MLPs) with two hidden layers, each containing 256 neurons. During training, the policy for step timing and region adjustments is running at 50 Hz, while the DCM trajectory generator and the whole-body controller run at 1 kHz.

We evaluate our method for the bipedal robot Kangaroo \cite{roigHardwareDesignControl2022} using the MuJoCo simulation environment \cite{todorovMuJoCoPhysicsEngine2012}. The Kangaroo robot is of height 145 cm with a mass of 40 kg. Each leg is fully actuated with 6 DoFs and has a flat foot. The feet measure 21 cm in length and 9 cm in width. We use PROX-QP \cite{bambadePROXQPAnotherQuadratic2022} to solve the QP for DCM trajectory generation and the whole-body controller. 

During training, push forces are randomly sampled from 200 N to 400 N and can act in any direction. Each disturbance is applied for 0.1 seconds, and each episode of reinforcement learning training lasts 20 seconds, with the robot experiencing a push every 1.5 seconds.

\subsection{Baseline method}
We adopt the model-based control approach, depicted in Fig. \ref{fig:overview}, as our baseline methodology. This baseline employs model-based step timing optimization (detailed in Section \ref{sec:dcm}) to adjust the remaining time $T_{\text{rem}}$. Within this framework, step adjustment for each foot occurs within a convex region on the same side, analogous to the scenario where $\theta=0^{\circ}$ in our approach.

\subsection{Quantitative evaluation on push recovery}
In this section, we evaluate our method of push recovery when walking forward with a reference velocity $v_{\text{ref}} = 0.3 \text{ms}^{-1}$. In all evaluations, the forces are applied for 0.1 seconds.

\begin{figure}
    \centering
    \includegraphics[width=0.45\textwidth]{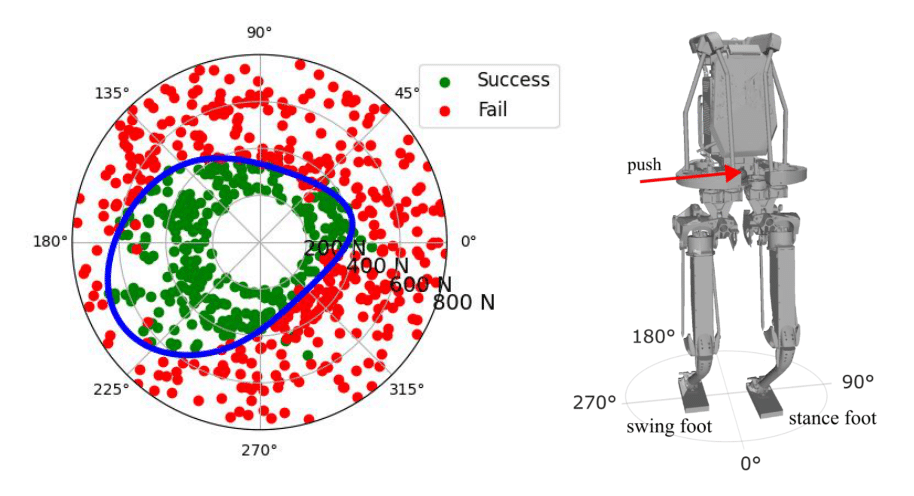}
    \caption{\textbf{Push force evaluation}. We conduct 1000 experiments to push the robot with random forces and in random directions. The SVM with RBF kernel is employed to obtain the maximal recoverable disturbance contour between successful and failed trials.}
    \label{fig:baseline}
    \vspace{-10pt}
\end{figure}
\subsubsection{Recovery force}
We push the robot at different times during a walking cycle. The push force ranges from 200N to 800N in any direction, which is up to two times more than the disturbance forces during training. The simulation is considered unsuccessful if the center of mass falls below a certain threshold.

Figure \ref{fig:baseline} presents the results of the simulations using our baseline method. We utilized a Support Vector Machine (SVM) with Radial Basis Function (RBF) kernels to classify successful and failed simulations, with the boundary representing the maximal recoverable disturbances. To comprehensively compare our method with the baseline, we introduce disturbances at different times during a robot's walking cycle. 

\begin{figure} []
    \centering
    \includegraphics[width=0.45\textwidth]{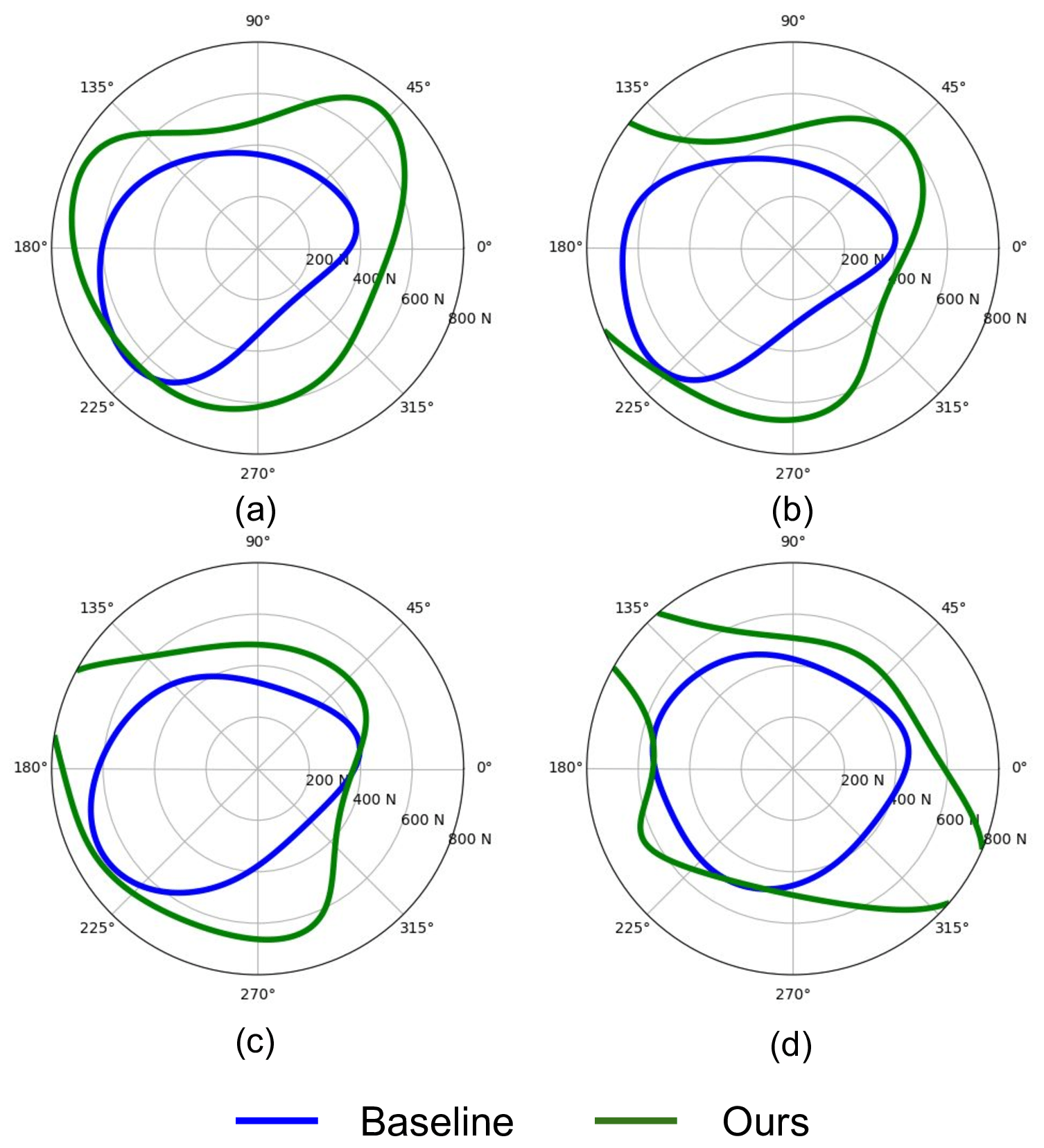}
    \caption{\textbf{Comparison of maximal recoverable disturbances}. Contours represent the maximal recoverable disturbance when a push is applied at different times during the transition phase. The push occurs at (a) the first half of the DS; (b) the second half of the DS; (c) the first half of the SS; (d) the second half of the SS. The robot is walking forward with a reference velocity of $v_{\text{ref}} = 0.3 \text{ms}^{-1}$ when the push occurs.}
    \label{fig:comparison}
    \vspace{-10pt}
\end{figure}

Figure \ref{fig:comparison}-(a) compares the maximal recoverable disturbances between the baseline and our methods when pushes occur during the first half of the double support phase. Fig. \ref{fig:comparison}-(b) shows the results for pushes occurring in the second half of the double support phase. Similarly, Fig. \ref{fig:comparison}-(c) and Fig. \ref{fig:comparison}-(d) illustrate the maximal recoverable disturbances for the first and second halves of the single support phase, respectively. Our RL-based footstep timing and footstep region adjustments significantly enhance walking robustness compared to the model-based step and timing adjustment in the baseline. Our method accommodates larger disturbance forces, regardless of the push direction and timing. As the robot is walking forward and the forward momentum of the robot partly compensates for the push, we observe a shift of all contour lines towards the backward direction ($180^{\circ}$). Furthermore, our proposed approach results in more noticeably asymmetric contour lines, indicating that RL improves recovery more effectively in certain directions. 

\subsection{Qualitative evaluation on push recovery}
\begin{figure*}
    \vspace{5pt}
    \centering
    \includegraphics[width=\linewidth]{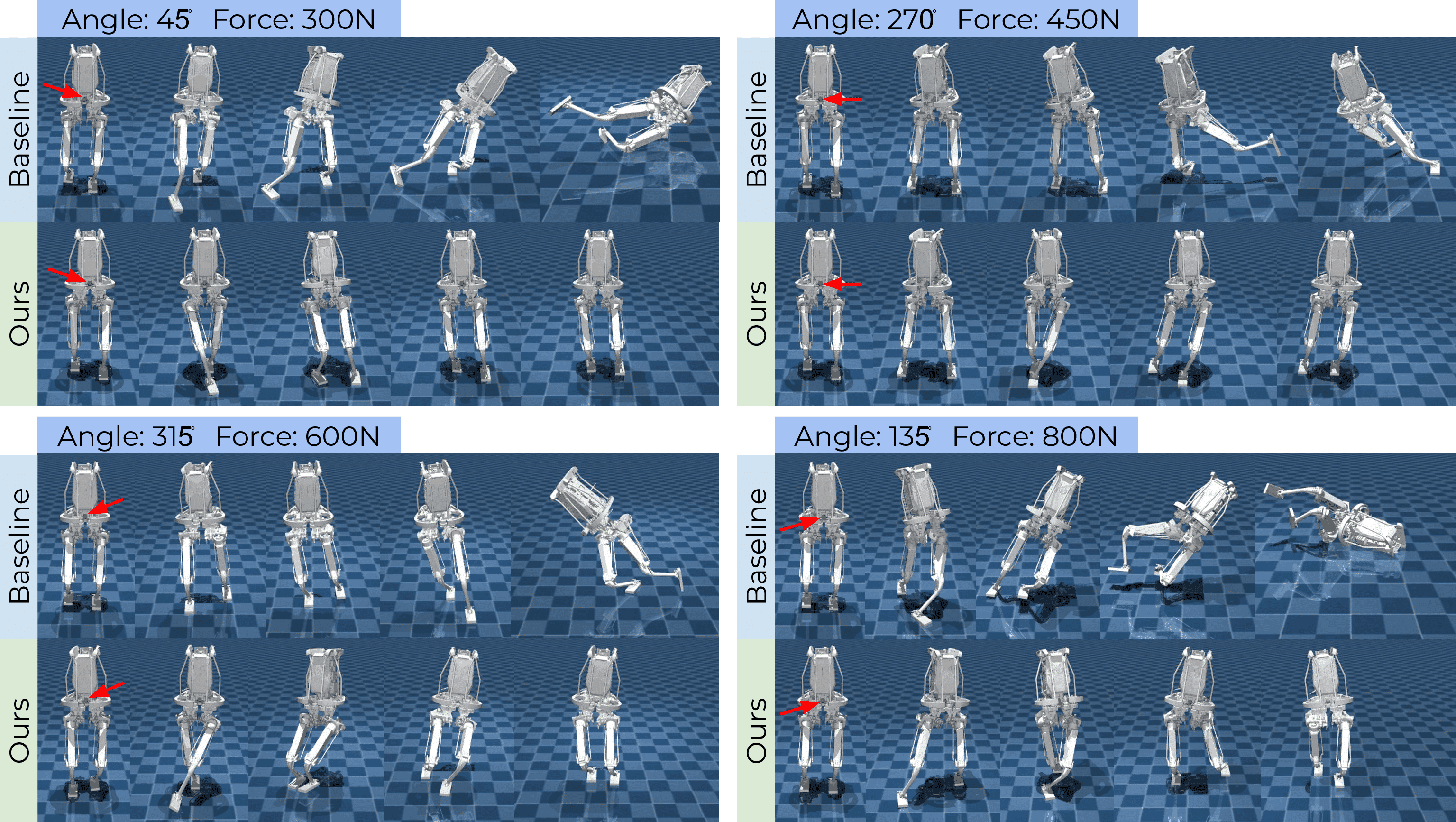}
    \caption{\textbf{Recovery from large disturbances}. We compare our method with the baseline under large disturbances, ranging from 300 N to 800 N. Using the baseline method, the robot falls in all four experiments. In contrast, our approach enables the robot to recover from all disturbances. Thanks to the step region adjustment, our approach allows the robot to cross its legs, effectively compensating for the large disturbances and facilitating a quick recovery.}
    \label{fig:qualitative}
    \vspace{-10pt}
\end{figure*}

To evaluate the advantages of foot region adjustment in our method, we push the robot with large forces while walking forward. Figure \ref{fig:qualitative} illustrates four experiments where the robot is pushed with different forces in various directions. Our approach recovers from the disturbances in all scenarios, whereas the baseline method results in the robot falling.


Thanks to our footstep region adjustment, the robot can place its foot in an effectively non-convex region, often leading to a cross-over leg behavior. Notably, in the last experiment shown in Fig. \ref{fig:qualitative}, the robot is subjected to a substantial force of 800N from the front right. Under the baseline method, the robot quickly falls due to the magnitude of the disturbance. In contrast, our method enables the robot to quickly cross over its legs to compensate for a significant portion of the momentum, leading to a rapid recovery from such a large disturbance.
\begin{figure*}
    \vspace{5pt}
    \centering
\includegraphics[width=\linewidth]{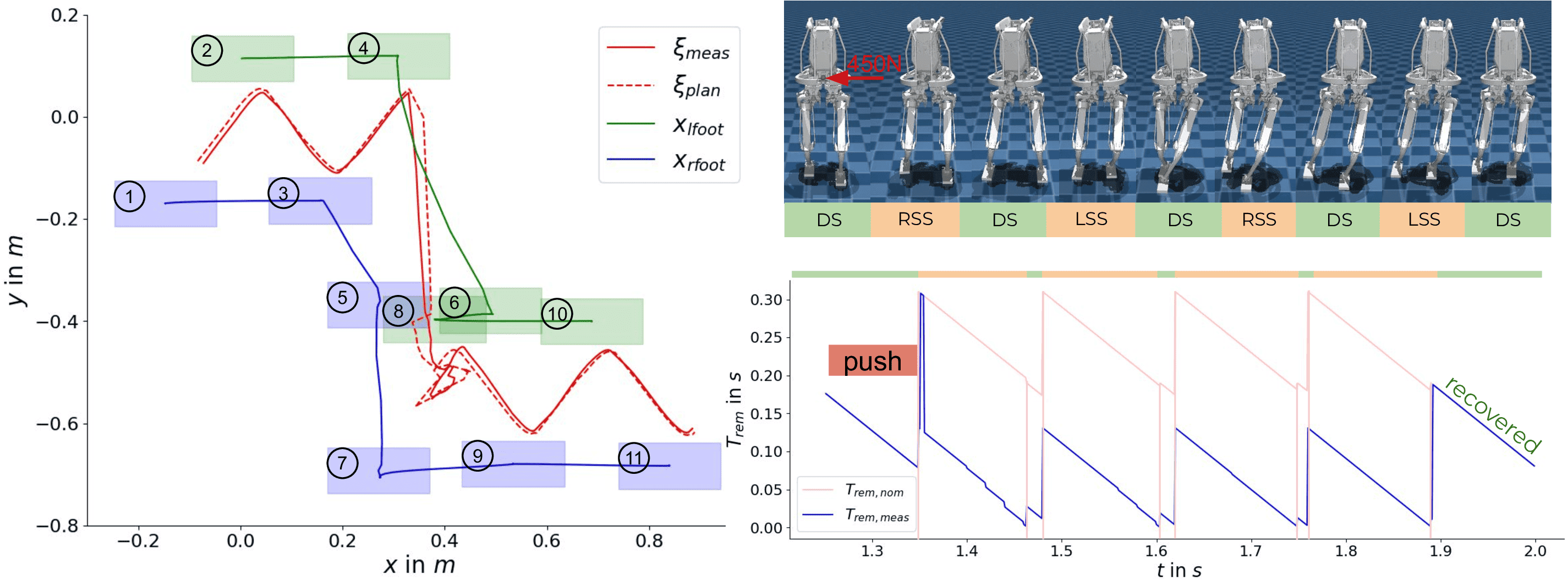}
    \caption{\textbf{Step adjustment in our approach}. The robot is pushed with 450N to the left. The left plot illustrates the adjusted footstep locations, showing a cross-over behavior at footsteps $\textcircled{5}$ and $\textcircled{6}$. Our method enhances recovery through rapid stepping by reducing the remaining time $T_{\text{rem}}$. Due to step timing and region adjustments, the robot recovers from the substantial push in approximately 0.55 seconds, deviating less than 0.5 meters from the desired trajectory. The top right snapshots correspond to the simulation in a MuJoCo environment (RSS: right single support; LSS: left single support).} 
    \label{fig:footplan}
    \vspace{-10pt}
\end{figure*}

Figure \ref{fig:footplan} illustrates the footstep adjustments when the robot is pushed to the left with a force of 450N. The left plot displays the footstep locations ($x_{\text{lfoot}}$ and $x_{\text{rfoot}}$) and the planned and measured DCM ( $\xi_{\text{plan}}$ and $\xi_{\text{meas}}$) during the simulation. To counteract the disturbance, the robot initially moves its right foot from position \textcircled{3} to \textcircled{5}, followed by crossing its left foot to the front right from position \textcircled{4} to \textcircled{6}. This sequence of movements significantly reduces the robot's momentum, enabling recovery in the subsequent step. The top right snapshots in Fig. \ref{fig:footplan} show the simulation results in a MuJoCo environment. The bottom right plot in Fig. \ref{fig:footplan} shows the nominal ($T_{\text{rem, nom}}$ in purple) and adjusted remaining time ($T_{\text{rem, meas}}$ in blue) during the double and single support phases. Our method effectively reduces $T_{\text{rem}}$, facilitating high-frequency stepping to maintain balance. Consequently, due to the combined adjustments in step timing and step region, the robot recovers from the substantial lateral push in approximately 0.55 seconds, deviating less than 0.5 meters from the nominal trajectory.

\section{Discussion and Conclusion}\label{sec:disc}
In this paper, we integrate reinforcement learning into a model-based control framework to address significant limitations in planning: footstep region and step timing. Unlike the model-based method that limits foot placement to a predefined convex region, we parametrize the step region with a convex region and a learnable parameter $\theta$. The RL learns to adjust the footstep region by rotating the convex region around the stance foot by the angle $\theta$. This allows the QP approach to find optimal foot positions that minimize the terminal DCM's distance to its target within an extended, effectively non-convex solution space. By focusing RL on adjusting the footstep region rather than precise footstep positions, we reduce the complexity for RL to learn, allowing for efficient training in a single environment on a CPU. 

In addition, we delegate the adjustment of step timing to RL, as optimizing step timing with a model-based method is challenging due to its nonlinearity. The time adjustment is managed by two parameters: step frequency $f$ and single support percentage $r$. The frequency $f$ dynamically changes the duration of a walking circle, while $r$ adjusts the time distribution between single and double support phases.

Our method combines data-driven RL with a model-based control framework by dynamically adjusting the parameters used in model-based optimization. This division of tasks leverages the strengths of both RL and model-based optimization, significantly enhancing the robot's walking robustness against external disturbances and improving the training efficiency of RL.

\bibliographystyle{IEEEtran}
\bibliography{IEEEabrv,IEEEsettings,relatedworks,HumanoidsRL24}

\begin{thebibliography}{10}
\providecommand{\url}[1]{#1}
\csname url@samestyle\endcsname
\providecommand{\newblock}{\relax}
\providecommand{\bibinfo}[2]{#2}
\providecommand{\BIBentrySTDinterwordspacing}{\spaceskip=0pt\relax}
\providecommand{\BIBentryALTinterwordstretchfactor}{4}
\providecommand{\BIBentryALTinterwordspacing}{\spaceskip=\fontdimen2\font plus
\BIBentryALTinterwordstretchfactor\fontdimen3\font minus \fontdimen4\font\relax}
\providecommand{\BIBforeignlanguage}[2]{{%
\expandafter\ifx\csname l@#1\endcsname\relax
\typeout{** WARNING: IEEEtran.bst: No hyphenation pattern has been}%
\typeout{** loaded for the language `#1'. Using the pattern for}%
\typeout{** the default language instead.}%
\else
\language=\csname l@#1\endcsname
\fi
#2}}
\providecommand{\BIBdecl}{\relax}
\BIBdecl

\bibitem{takenakaRealTimeMotion2009a}
T.~Takenaka, T.~Matsumoto, and T.~Yoshiike, ``Real time motion generation and control for biped robot -1st report: {{Walking}} gait pattern generation-,'' in \emph{Proc. {{IEEE}}/{{RSJ Int}}. {{Conf}}. {{Intell}}. {{Robots Syst}}.}, 2009.

\bibitem{englsbergerThreeDimensionalBipedalWalking2015}
J.~Englsberger, C.~Ott, and A.~{Albu-Sch{\"a}ffer}, ``Three-{{Dimensional Bipedal Walking Control Based}} on {{Divergent Component}} of {{Motion}},'' \emph{IEEE Transactions on Robotics}, 2015.

\bibitem{mesesanOnlineDCMTrajectory2021}
G.~Mesesan, J.~Englsberger, and C.~Ott, ``Online {{DCM Trajectory Adaptation}} for {{Push}} and {{Stumble Recovery}} during {{Humanoid Locomotion}},'' in \emph{Proc. {{IEEE Int}}. {{Conf}}. {{Robot}}. {{Automat}}.}, 2021.

\bibitem{Khazoom_2024}
C.~Khazoom, S.~Hong, M.~Chignoli, E.~Stanger-Jones, and S.~Kim, ``Tailoring solution accuracy for fast whole-body model predictive control of legged robots,'' \emph{IEEE Robotics and Automation Letters}, 2024.

\bibitem{khadivWalkingControlBased2020}
M.~Khadiv, A.~Herzog, S.~{\relax Ali}.~A. Moosavian, and L.~Righetti, ``Walking {{Control Based}} on {{Step Timing Adaptation}},'' \emph{IEEE Transactions on Robotics}, 2020.

\bibitem{egleStepTimingAdaptation2023}
T.~Egle, J.~Englsberger, and C.~Ott, ``Step and {{Timing Adaptation}} during {{Online DCM Trajectory Generation}} for {{Robust Humanoid Walking}} with {{Double Support Phases}},'' in \emph{{{IEEE-RAS}} 22nd {{Int}}. {{Conf}}. {{Humanoid Robots}}}, 2023.

\bibitem{yangLearningWholebodyMotor2018}
C.~Yang, K.~Yuan, W.~Merkt, T.~Komura, S.~Vijayakumar, and Z.~Li, ``Learning {{Whole-body Motor Skills}} for {{Humanoids}},'' in \emph{{{IEEE-RAS}} 18th {{Int}}. {{Conf}}. {{Humanoid Robots}}}, Nov. 2018.

\bibitem{ferigoEmergenceWholeBodyStrategies2021}
D.~Ferigo, R.~Camoriano, P.~M. Viceconte, D.~Calandriello, S.~Traversaro, L.~Rosasco, and D.~Pucci, ``On the {{Emergence}} of {{Whole-Body Strategies From Humanoid Robot Push-Recovery Learning}},'' \emph{IEEE Robotics and Automation Letters}, 2021.

\bibitem{horakCentralProgrammingPostural1986}
F.~B. Horak and L.~M. Nashner, ``Central programming of postural movements: Adaptation to altered support-surface configurations,'' \emph{Journal of Neurophysiology}, no.~6, 1986.

\bibitem{englsbergerTorqueBasedDynamicWalking2018a}
J.~Englsberger, G.~Mesesan, A.~Werner, and C.~Ott, ``Torque-{{Based Dynamic Walking}} - {{A Long Way}} from {{Simulation}} to {{Experiment}},'' in \emph{Proc. {{IEEE Int}}. {{Conf}}. {{Robot}}. {{Automat}}.}, 2018.

\bibitem{singhLearningBipedalWalking2023}
R.~P. Singh, Z.~Xie, P.~Gergondet, and F.~Kanehiro, ``Learning {{Bipedal Walking}} for {{Humanoids}} with {{Current Feedback}},'' \emph{IEEE Access}, 2023.

\bibitem{xieALLSTEPSCurriculumdrivenLearning2020a}
Z.~Xie, H.~Y. Ling, N.~H. Kim, and M.~{van de Panne}, ``{{ALLSTEPS}}: {{Curriculum-driven Learning}} of {{Stepping Stone Skills}},'' \emph{Computer Graphics Forum}, 2020.

\bibitem{rudinLearningWalkMinutes2022}
N.~Rudin, D.~Hoeller, P.~Reist, and M.~Hutter, ``Learning to {{Walk}} in {{Minutes Using Massively Parallel Deep Reinforcement Learning}},'' in \emph{Proceedings of the 5th {{Conference}} on {{Robot Learning}}}, 2022.

\bibitem{liReinforcementLearningVersatile2024}
Z.~Li, X.~B. Peng, P.~Abbeel, S.~Levine, G.~Berseth, and K.~Sreenath, ``Reinforcement {{Learning}} for {{Versatile}}, {{Dynamic}}, and {{Robust Bipedal Locomotion Control}},'' arXiv, Tech. Rep. arXiv:2401.16889, Jan. 2024.

\bibitem{siekmannBlindBipedalStair2021}
J.~Siekmann, K.~Green, J.~Warila, A.~Fern, and J.~Hurst, ``Blind {{Bipedal Stair Traversal}} via {{Sim-to-Real Reinforcement Learning}},'' arXiv, Tech. Rep. arXiv:2105.08328, May 2021.

\bibitem{duanLearningDynamicBipedal2022}
H.~Duan, A.~Malik, M.~S. Gadde, J.~Dao, A.~Fern, and J.~Hurst, ``Learning {{Dynamic Bipedal Walking Across Stepping Stones}},'' in \emph{2022 {{IEEE}}/{{RSJ International Conference}} on {{Intelligent Robots}} and {{Systems}} ({{IROS}})}, Oct. 2022.

\bibitem{baoDeepReinforcementLearning2024}
L.~Bao, J.~Humphreys, T.~Peng, and C.~Zhou, ``Deep {{Reinforcement Learning}} for {{Bipedal Locomotion}}: {{A Brief Survey}},'' arXiv, Tech. Rep. arXiv:2404.17070, Apr. 2024.

\bibitem{gu2023walkingbylogicsignaltemporallogicguided}
Z.~Gu, R.~Guo, W.~Yates, Y.~Chen, and Y.~Zhao, ``Walking-by-logic: Signal temporal logic-guided model predictive control for bipedal locomotion resilient to external perturbations,'' 2023.

\bibitem{duanLearningTaskSpace2021}
H.~Duan, J.~Dao, K.~Green, T.~Apgar, A.~Fern, and J.~Hurst, ``Learning {{Task Space Actions}} for {{Bipedal Locomotion}},'' in \emph{2021 {{IEEE International Conference}} on {{Robotics}} and {{Automation}} ({{ICRA}})}, May 2021.

\bibitem{castilloReinforcementLearningBasedCascade2022}
G.~A. Castillo, B.~Weng, W.~Zhang, and A.~Hereid, ``Reinforcement {{Learning-Based Cascade Motion Policy Design}} for {{Robust 3D Bipedal Locomotion}},'' \emph{IEEE Access}, vol.~10, 2022.

\bibitem{castilloTemplateModelInspired2023b}
G.~A. Castillo, B.~Weng, S.~Yang, W.~Zhang, and A.~Hereid, ``Template {{Model Inspired Task Space Learning}} for {{Robust Bipedal Locomotion}},'' arXiv, Tech. Rep. arXiv:2309.15442, Sep. 2023.

\bibitem{griffinReachabilityAwareCapture2023}
R.~Griffin, J.~Foster, S.~Fasano, B.~Shrewsbury, and S.~Bertrand, ``Reachability {{Aware Capture Regions}} with {{Time Adjustment}} and {{Cross-Over}} for {{Step Recovery}},'' arXiv, Tech. Rep. arXiv:2307.11968, Jul. 2023.

\bibitem{habibHandlingNonConvexConstraints2022}
A.~S. Habib, F.~M. Smaldone, N.~Scianca, L.~Lanari, and G.~Oriolo, ``Handling {{Non-Convex Constraints}} in {{MPC-Based Humanoid Gait Generation}},'' in \emph{2022 {{IEEE}}/{{RSJ International Conference}} on {{Intelligent Robots}} and {{Systems}} ({{IROS}})}, Oct. 2022.

\bibitem{mesesanConvexPropertiesCenterofMass2018}
G.~Mesesan, J.~Englsberger, C.~Ott, and A.~{Albu-Sch{\"a}ffer}, ``Convex {{Properties}} of {{Center-of-Mass Trajectories}} for {{Locomotion Based}} on {{Divergent Component}} of {{Motion}},'' \emph{IEEE Robotics and Automation Letters}, vol.~3, no.~4, pp. 3449--3456, 2018.

\bibitem{schulmanProximalPolicyOptimization2017}
J.~Schulman, F.~Wolski, P.~Dhariwal, A.~Radford, and O.~Klimov, ``Proximal {{Policy Optimization Algorithms}},'' arXiv, Tech. Rep. arXiv:1707.06347, Aug. 2017.

\bibitem{roigHardwareDesignControl2022}
A.~Roig, S.~K. Kothakota, N.~Miguel, P.~Fernbach, E.~M. Hoffman, and L.~Marchionni, ``On the {{Hardware Design}} and {{Control Architecture}} of the {{Humanoid Robot Kangaroo}},'' in \emph{6th {{Workshop}} on {{Legged Robots}} during the {{Int}}. {{Conf}}. {{Robot}}. {{Automat}}.}, 2022.

\bibitem{todorovMuJoCoPhysicsEngine2012}
E.~Todorov, T.~Erez, and Y.~Tassa, ``{{MuJoCo}}: {{A}} physics engine for model-based control,'' in \emph{Proc. {{IEEE}}/{{RSJ Int}}. {{Conf}}. {{Intell}}. {{Robots Syst}}.}, 2012, pp. 5026--5033.

\bibitem{bambadePROXQPAnotherQuadratic2022}
A.~Bambade, S.~{El-Kazdadi}, A.~Taylor, and J.~Carpentier, ``{{PROX-QP}}: {{Yet}} another {{Quadratic Programming Solver}} for {{Robotics}} and beyond,'' in \emph{Robotics: {{Science}} and {{Systems XVIII}}}, Jun. 2022.

\end{thebibliography}

\end{document}